\documentclass[AMA,STIX1COL]{WileyNJD-v2}
\usepackage[super]{NJDnatbib}
\usepackage{moreverb}
\usepackage{subcaption}

\usepackage{graphicx}

\usepackage{xcolor, soul}
\sethlcolor{pink}  
\usepackage{algorithm}%
\usepackage{algpseudocode}

\newcommand\BibTeX{{\rmfamily B\kern-.05em \textsc{i\kern-.025em b}\kern-.08em
T\kern-.1667em\lower.7ex\hbox{E}\kern-.125emX}}

\articletype{Original Article}%

\received{<day> <Month>, <year>}
\revised{<day> <Month>, <year>}
\accepted{<day> <Month>, <year>}

\begin{document}

\title{
Plug-in for visualizing 3D tool tracking from videos of Minimally Invasive surgeries}

\author[1]{Shubhangi Nema*}
\author[2]{Abhishek Mathur}

\author[1]{Prof. Leena Vachhani}

\authormark{Nema \textsc{et al}}

\address{Shubhangi and  Prof. Leena are with \orgdiv{Systems and Control Group}, \orgname{Indian Institute of Technology Bombay}, \orgaddress{\state{Maharashtra}, \country{India} }(e-mail: Shubhangi Nema, 194234001@iitb.ac.in; Leena Vachhani, leena.vachhani@iitb.ac.in)}

\address{Abhishek is with \orgdiv{Mechanical Engineering}, \orgname{Indian Institute of Technology Goa}, \orgaddress{\state{Goa}, \country{India} }(e-mail: Abhishek Mathur, abhishekramanmathur6@gmail.com)}

\corres{*Shubhangi Nema, Ph.D. Research Scholar, Systems and Control Group, Indian Institute of Technology Bombay, Mumbai, 470001, India. \email{194234001@iitb.ac.in}
\newline \newline \textbf{Funding} \newline Financial support for this study was provided by Prime Minister's Research Fellows (PMRF) scheme, India (PMRF Id no. 1300229 dated May 2019) for pursuing research in higher educational institutions in India. \url{https://www.pmrf.in/}. 
}

\abstract[Abstract]{

\textbf{Background} 
\newline This paper tackles instrument tracking and 3D visualization challenges in minimally invasive surgery (MIS), crucial for computer-assisted interventions. Conventional and robot-assisted MIS encounter issues with limited 2D camera projections and minimal hardware integration. 

\noindent\textbf{Method}
\newline The objective is to track and visualize the entire surgical instrument, including shaft and metallic clasper, enabling safe navigation within the surgical environment. The proposed method involves 2D tracking based on segmentation maps, facilitating creation of labeled dataset without extensive ground-truth knowledge. Geometric changes in 2D intervals express motion, and kinematics based algorithms process results into 3D tracking information.

\noindent\textbf{Result} 
\newline 
Synthesized and experimental results in 2D and 3D motion estimates demonstrate negligible errors, validating the method for labeling and motion tracking of instruments in MIS videos.

\noindent\textbf{Conclusion}
\newline
The conclusion underscores the proposed 2D segmentation technique's simplicity and computational efficiency, emphasizing its potential as direct plug-in for 3D visualization in instrument tracking and MIS practices.
\newline 
}

\keywords{Minimally Invasive Surgery, Surgical instruments, 3D tracking, Geometric cues}

\maketitle

\section{Introduction}

Advances in robotics, computer graphics, and virtual reality (VR) have been increasingly applied to surgical robotics \citep{R1}. In the cutting-edge interdisciplinary research field of information and medical sciences, research on virtual surgery simulation systems has significant application value for reducing surgery risks, cutting training costs, and protecting human health \citep{R2, R3}. With the help of virtual surgery training platform, trainee surgeons can skillfully master the operations of surgical instruments, the general procedure of surgery, and the anatomy of a diseased region or organ. 
The accurate 3-dimensional (3D) movement tracking of surgical instruments is a vital part of the VR-based surgery training simulation system 
\citep{beulens2020analysis,nema2022surgical}. Assessment of surgical instrument motion can be implemented to objectively classify skill levels.
In robotic surgery, surgeons can view the surgical scene on a screen in the form of 2D projections as captured by the camera \citep{dataset}. Segmentation of surgical instruments used during surgery \citep{Pakhomov, Shvets, nema2023unpaired} is a first step to tracking their motions for understanding the surgical procedure. We aim to develop techniques for 3D motion tracking of MIS instruments from raw surgical videos available currently. The multi-class segmentation problem \citep{nema2023unpaired} has been formulated to distinguish between different instruments or different parts of an instrument (connected sets of a surgical instrument) from the background. 
A deep residual learning-based technique \citep{Pakhomov} incorporates dilated convolutions for multi-class segmentation by advancing binary-segmentation performance. Advancements in multi-class segmentation of robotics and rigid instruments \citep{nema2023unpaired} using unpaired training approach conveys feasibility of good accuracy segmentation on unseen data as well. 
The segmentation maps are readily available from these methods. We investigate the segmented maps to track the 2D movements of the surgical instruments used to carry out any interventions. These movements can not be directly translated into real-world scenarios; thus, a 3D tracking tool is necessary to have better scene perception and carry out smooth movements of instruments by the surgeons.

The 3D motion of surgical instrument has been attempted by either making physical modifications to have good image contrast or having some knowledge of the instrument or environment in captured raw surgical videos.

\underline{Methods based on physical modification:}
To track instrument motion, the modifications are done to improve the appearance of instrument in the captured video.
Typically markers have been placed on instruments and tracking of markers renders the 3D motion of instruments. Work by Zhang \textit{et al.} \citep{zhang2002application} uses the \textit{intensity information} of the image for the surgical instrument marker localization. The evaluation consists of several experiments such as camera calibration, endoscope image distortion correction, endoscope holder simulation, and visual tracking using pcBird.  
There has also been the use of 3D trackers in virtual reality. According to their physical properties, they are roughly classified into five subcategories: mechanical tracker \citep{R4}, magnetic tracker \citep{R5}, ultrasonic tracker \citep{zhang2017real,R6}, optical tracker \citep{R7} and hybrid tracker \citep{R8}. Some of them can provide high positioning accuracy \citep{R9}, and have been used in medical applications. 
A 3D surgical instrument tracking and positioning method with a high performance-price ratio has been highly desirable for computer-based virtual surgery simulation systems \citep{R11}. An iterative approach \citep{seslija2008feasibility} based on the projection-Procrustes technique determines the three-dimensional positions and orientations of known sparse objects from a single, perspective projection. 
A 3D tracking and positioning method for surgical instruments \citep{duan20113d} based on \textit{stereoscopic vision} combines force sensor and embedded acquisition device is proposedin order to measure soft tissues parameters. 
An algorithm for real-time tracking of laparoscopy instruments \citep{gautier2021real} \textit{ attached with colored tapes (markers)} uses the video cues of a standard physical laparoscopy training box with a single fisheye camera. An intelligent tracking system for surgical instruments is proposed based on data fusion of \textit{multicamera modules} \citep{chen2023intelligent}. The method evaluated for static positioning and dynamic trajectory tracking.
However, these existing devices do not provide cost-effective solutions limiting their usage by medical centers and research institutes.
Moreover, there are practical difficulties in adopting the use of external markers during
surgery. The use of markers increases the procedural time during surgery and needs an ethics board's approval. These are accompanied by the risk of infection and occlusion  of markers affecting tracking performance.

\underline{Methods using raw surgical videos:}
As markers or contrasting colored surgical instruments require changing procedures, the investigations have attempted to directly use raw surgical videos without making any procedural changes.
A 2D tooltip location tracking algorithm in laparoscopic training videos analyses hand \& eye coordination \citep{jiang2015video}. The tool-tip detection algorithm is based on background subtraction and thresholding foreground showing promising outcomes in occlusion-free and well-contrasting surgical scenarios. 
The geometric and photometric invariants common to standard Fundamentals of Laparoscopic Surgery (FLS) training boxes have been  utilized \citep{allen2011visual} for tracking the spatial motion of standard laparoscopic instruments from videos. 

Early image-based methods predominately estimate instrument poses based on image processing techniques that utilize hand-crafted visual features (sometimes apriori known) and more complex learned discriminative models. Tracking of an instrument is performed independently on each frame, and the methods are typically fast and robust, and handling complex and fast motion becomes easier. 
A 2D tracker based on a \textit{Generalized Hough Transform using SIFT features} is developed \citep{du2016combined} for tracking surgical instruments for minimally invasive and robotic-assisted surgery. The technique initializes the 3D tracker at each frame which recovers the 3D instrument pose over long sequences. A few direct 3D tracking based on deep learning methods  have also been investigated. 
A fully convolutional network (FCN) with optical flow for tracking \citep{garcia2016real} and TernausNet-16, a deep convolutional neural network-based instrument tracking \citep{cheng2021deep}; an extension of U-Net \citep{ronneberger2015u} constructed on VGG-16 \citep{simonyan2014very} are two classic examples. 
However, these methods are computationally demanding and necessitate a large volume of labeled datasets.

Our approach is to disintegrate the problem of 3D tracking through 2D rendered images by identifying the 2D movement of the surgical instrument and then converting it to the 3D motion using geometric dynamics. Moreover, this work aims to track each part of the instrument for avoiding collision with the sensitive tissues/vessels.
The objective is to use lower dimensional results to track instruments in higher dimensions with simple geometric computations for visualization of 3D instrument motion and thus facilitating creation of labeled dataset (\textit{in terms of motion labels for position and rotation angles with respect to time}).

A surgical instrument has mainly two parts, the shaft, and the metallic clasper/end-effector. The metallic clasper attached to the shaft performs critical interventions such as grasping, cutting, etc.  
We use interval arithmetic approach \citep{nema2021safe}to consider the complete shape of the instrument in terms of intervals to avoid collision of any part of the instrument with critical organs. 

The summary of contributions of the proposed work building up the structure of the rest of this paper is as follows:
\begin{itemize}
    \item 
   A novel approach that combines a unique methodology and problem formulation for 2D tracking in surgical instruments is presented next. It employs kinematics to process the 2D tracking results and generate corresponding 3D movements. The approach considers the instrument's shape through connected intervals, analyzes geometric changes within these intervals to estimate motion, and results in an algorithm that accommodates various shapes while enhancing tool dexterity.
  
    \item The proposed algorithms for 2D and 3D tracking of a surgical instrument for a complete surgery are presented in Section 3. The algorithm uses a sequence of simple functions from an existing open-source vision package and computes the 3D action of instrument parts (shaft and clasper).
    \item Section 4 presents simulations to visualize the 3D motion of surgical instruments. The results from these simulations provide evidence of the precise and accurate movements achieved by the instruments.
    The error analysis in 2D and 3D is performed to assess the accuracy claims, that eliminates the necessity to rely on ground truth information. The quantitative experimental results for 3D tracking are compared to state-of-the-art methods, showing a significant reduction in error. Moreover, the average error obtained from the proposed method using monocular vision is comparable to that of stereo vision-based 3D tracking. The proposed approach produces tracking results without any physical intervention and can be used for autonomous tracking of surgical instruments during MIS.
    The future scope of this work to implement the strategy in an experimental setup for providing real-time training or assistance to the surgeons is discussed in Section 5, along with conclusions.  
\end{itemize}

\section{Proposed work}

\begin{figure*}
\centerline{\includegraphics[width=\linewidth,height=5cm]{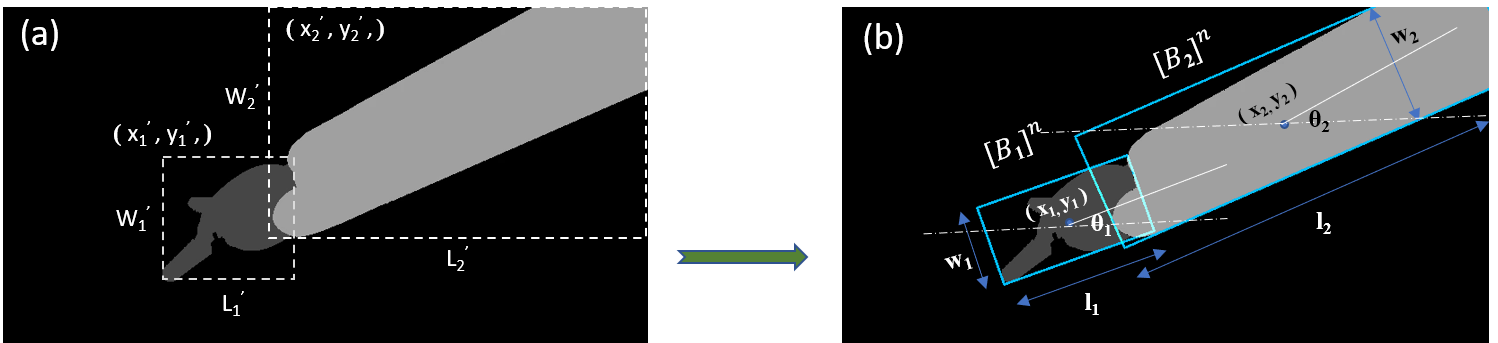}}
\caption{Surgical instrument representation (\textit{zoomed view}) with two links (k= \{1,2\}) for nth frame
(a) Parameters of $R_k^n$ (b) Parameters of $[B_k]^n$}
\label{fig6}
\end{figure*}

In an image frame of endoscopic video (frame size of (720 $\times$ 526 $pixel^2$) in the dataset\citep{dataset}), we propose to represent the surgical instrument as a combination of approximated rectangular intervals. Let the length of the video be $N$ image frames and the number of instrument parts is $K$. The approximated rectangular interval  corresponding to each instrument part ($k=1,\ldots, K$) in each $n^{th}$ image frame ($I^n$, $n \in [1, N]$) is $[B_k]^n := \{x_k,y_k, l_k, w_k, \theta_k\}$, with center ($x_k,y_k$), length ($l_k$), width ($w_k$), angle ($\theta_k$) (\textit{refer Fig. \ref{fig6} (a)}). The angle $\theta_k$ is with respect to the axis aligned with respect to the $k-1$ interval in the 2D projected image of the instrument. 
The instrument configuration comprises horizontal , vertical positions and rotation on the image plane. This rotation represents the rotation of the projected instrument (on image plane) separately for each part of the instrument. We consider the image plane as $X-Y$ plane, with $Z-$ axis orthogonal and projecting out of the image plane. A surgical instrument in an image projection in the proposed work has two connected intervals defined using bounding boxes for the metallic clasper ($[B_1]^n, k=1$) and the shaft ($[B_2]^n, k=2$), an illustration of the same is in Fig. \ref{fig6}(b). 

The overlapping region between two connected intervals is considered in this work to incorporate actual movement between the shaft and the clasper through a linkage.
The interval coordinates or bounding boxes ($[B_k]^n$ with center ($x_k,y_k$), length($l_k$), width($w_k$), angle($\theta_k$) surrounding the metallic clasper ($k=1$) and the shaft ($k=2$) along with their corresponding corner points are required to track 3D motion for surgical instruments. The 2D segmented images ($I^n| n= 1,\ldots, N$) of the surgical instruments are taken as inputs in order to generate these bounding boxes using Algorithm \ref{algo:boundingbox}. 
Each bounding box in an image frame $n$ has four vertices  or corners $\{(x_{k,i}^n,y_{k,i}^n)|_ {i= \{1,2,3,4\}}\}$ which serve to be the primary input that can be processed using Algorithm \ref{algo:3D track} 
The corner points for a bounding box are obtained using the description of $[B_k]^n$ as follows: 
\begin{eqnarray}
(x_{k,1}^n,y_{k,1}^n)=
     (x_k-(l_k/2) cos\theta_k, y_k + (w_k/2) sin\theta_k) \nonumber\\
 (x_{k,2}^n,y_{k,2}^n)=   (x_k+(l_k/2) cos\theta_k, y_k+(w_k/2) sin\theta_k)\\
  (x_{k,3}^n,y_{k,3}^n)= (x_k-(l_k/2) cos\theta_k, y_k-(w_k/2) sin\theta_k)\nonumber\\
 (x_{k,4}^n,y_{k,4}^n)=  (x_k + (l_k/2) cos\theta_k, y_k-(w_k/2) sin\theta_k).\nonumber
\end{eqnarray}

The 3D movements for the instrument's clasper and the shaft are generated in the form of translation matrix ($\mathcal{T}_k$) and rotation matrix ($\mathcal{R}_k$) using the following steps:
Let $\Delta x_k$, $\Delta y_k$, and $\Delta z_k$ abe the movement of the instrument's $k^{th}$ part ($k=1$ for clasper and $k=2$ for shaft)  in $x$, $y$ and $z$ direction respectively.  
The proposed method, as its first step, tracks the motion of the centroid of the bounding box.
In particular, change in the $x-$coordinate of the centroid ($C_{k,x}$) gives $\Delta x_k$ and change in the $y-$coordinate of the centroid ($C_{k,y}$) gives y ($\Delta y_k$).

The centroids ($C_{k,x}^n$ and $C_{k,y}^n$) of the bounding box are given by
\begin{equation} 
\label{eq 6}
C_{k,x}^n = \sum_{i=1}^{4} x_{k,i}^n /4 , 
C_{k,y}^n = \sum_{i=1}^{4} y_{k,i}^n /4
\end{equation}

Now to compute the movement in  $z-$ direction ($\Delta z_k$), we capture the geometric changes in the bounding box $[B_k]^n$.
 The area of a bounding box is given by \footnote{\% is the modulo operator giving the remainder of the division operation}.
\begin{equation}
\label{eq 1}
A_k^n =  1/2 *(\sum_{i=1}^{3}(x_{k,i}^n*y_{k,(i\%4+1))}^n-\sum_{i=1}^{3}(x_{k,(i\%4+1)}^n*y_{k,i}^n)) 
\end{equation} 

This is required to compute the change in scale ($S_k^n$) in each frame which is evaluated by taking the ratio of areas of bounding boxes in the current ($n$) and previous frame ($n-1$) respectively.
Thus, the $z-$ coordinates for a bounding box is evaluated using the  scale change ($S_k^n$), area ($A_k^n$), and the given focal length $f$ of the endoscope as follows:
\begin{equation}
\label{eq5}
z_k^n \approx sqrt(f^2 . (S_k^n)^2 / A_k^n)
\end{equation}

Now, using $\Delta z_k = z_k^n- z_k^{n-1}$, the translation matrix $\mathcal{T}_k$ is given by
\begin{equation}
\label{eq 7}
\mathcal{T}_k \approx  T_Z(\Delta z_k) T_Y( \Delta y_k) T_X( \Delta x_k)
\end{equation}
\newline
where,
$$
T_Z(\Delta z_k)
\approx
\begin{bmatrix}
1 & 0 & 0\\
0 & 1 & 0\\
0 & 0 & \Delta z_k
\end{bmatrix}
$$
$$
T_Y( \Delta y_k)
=
\begin{bmatrix}
1 & 0 & 0\\
0 & 1 & \Delta y_k \\
0 & 0 & 1
\end{bmatrix}
$$

$$
T_X(\Delta x_k)
=
\begin{bmatrix}
1 & 0 & \Delta x_k\\
0 & 1 & 0 \\
0 & 0 & 1
\end{bmatrix}
$$

For computing the rotation matrix $\mathcal{R}_k$, the net change in roll, pitch, and yaw orientations of each instrument part, $ (\Delta  \phi_k)$, $ (\Delta  \theta_k)$, and $\Delta  \psi_k)$ respectively is computed using the changes in internal angle of each edge of the bounding box. The slope of each edge of the bounding box is given by \begin{equation}
\label{eq 4}
m_{k,i}^n = (y_{k,i\%4+1}^n - y_{k,i}^n) / (x_{k,i\%4+1}^n - x_{k,i}^n),
\end{equation}
and each internal angle is obtained by
$a_{k,i}^n$ is defined as,
\begin{equation}
\label{eq 2}
a_{k,i}^n = tan^{-1}((m_{k,{i\%4+1}}^n - m_{k,i}^n)/(1+m_{k,i\%4+1}^n*m_{k,i}^n))
\end{equation}

The change in roll-pitch-yaw are computed by
capturing the following geometric changes heuristically:
\begin{itemize}
\item $\Delta  \phi_k$ is the rotation about the $x-$axis. The projection of the bounding box on the  image plane changes the internal angle of  the bounding box. Hence, the change in  roll is captured as the difference between the individual sums of $\angle 1$ ($a_{k,1}^n$) $\&$ $\angle 2$ ($a_{k,2}^n$) and $\angle 3$ ($a_{k,3}^n$) $\&$ $\angle 4$ ($a_{k,4}^n$). The change in angle  is zero when the instrument part is parallel to the image plane, but it develops a non-zero value directly proportional to the roll of the corresponding instrument part.
\item $\Delta  \theta_k$ is the rotation about the y axis and  the change in pitch is captured as the difference between the individual sums of $\angle 1$ ($a_{k,1}^n$) $\&$ $\angle 4$ ($a_{k,4}^n$) and $\angle 3$ ($a_{k,3}^n$) $\&$ $\angle 2$ ($a_{k,2}^n$). 
\item In the case of $ \Delta  \psi_k$, the axis is orthogonally  out of the image plane. Hence, we find the initial inclination (defined using $m_{k,i}^n$) of the sides of the bounding box and subtract the average from $\pi$ radians. 
\end{itemize}

Thus, the rotation matrix $\mathcal{R}_k$ matrix is computed using 
\begin{equation}
\label{eq 3}
\mathcal{R}_k \approx  R_Z(\Delta \psi_k) R_Y( \Delta \theta_k) R_X( \Delta \phi_k)
\end{equation}
\newline
where,
$$
R_Z(\Delta \psi_k)
=
\begin{bmatrix}
cos( \Delta \psi_k) & -sin( \Delta \psi_k) & 0\\
sin( \Delta \psi_k) & cos( \Delta \psi_k) & 0\\
0 & 0 & 1
\end{bmatrix}
$$
$$
R_Y( \Delta \theta_k)
=
\begin{bmatrix}
cos(\Delta \theta_k) & 0 & sin( \Delta \theta_k)\\
0 & 1 & 0 \\
-sin(\Delta \theta_k) & 0 & cos(\Delta \theta_k)
\end{bmatrix}
$$

$$
R_X(\Delta \phi_k)
=
\begin{bmatrix}
1 & 0 & 0\\
0 & cos(\Delta \phi_k) & -sin(\Delta \phi_k) \\
0 & sin(\Delta \phi_k) & cos(\Delta \phi_k)
\end{bmatrix}
$$

A novel 3D motion tracking algorithm to robustly estimate the full 3D pose of surgical instruments in minimally invasive surgery using 2D image projections is presented next.

\begin{algorithm}

\textbf{Result:}{$\{x_{k,i}^n,y_{k,i}^n \} |$ k=1,2 and i=1,2,3,4}

\textbf{Input:} { $I^n$} 
\begin{algorithmic}[1]
 \For{$k = \{1,2\}$:}
  
\State  $M_k^n = mask(I^n, \mbox{intensity value range for box } k)$;

 \State $O_k^n$ = $I^n$ \& $M_k^n$;

 \State $G_k^n = gray(O_k^n)$;

 \State $H_k^n = contour$( $G_k^n$, contour retrieval Mode, technique);

\State  $R_k^n$ = $rect\_bound(O_k^n, H_k^n)$;

 \State $[B_k^n] = minAreaRect(R_k^n)$;

\State  $x_{k,i}^n,y_{k,i}^n = boxPoints(B_k^n)$;
 
\EndFor
\end{algorithmic}
\caption{Bounding boxes for a surgical instrument in an image frame: $Bound\_box$ ($I_n$)}
\label{algo:boundingbox}
\end{algorithm}

\section{Algorithms}

This section demonstrates the Algorithm \ref{algo:boundingbox} and \ref{algo:3D track} based on geometric changes in intervals and kinematics.

The $n^{th}$ image frame ($I^n$)  of the video sequence, where $n=1,\ldots, N$ contains three regions, the metallic clasper, shaft and the background represented using different pixel values.

Algorithm~\ref{algo:boundingbox} generates the bounding boxes for the clasper ($[B_1^n], k=1$) and the shaft ($[B_2^n], k=2$) and their respective vertices or bounding box points ($(x_{k,i}^n,y_{k,i}^n)$ from the image frame $I^n$. The bounding box interval and the corresponding internal angles changes with each image frame during $z$ movement of the individual instrument part.
The metallic clasper and the shaft are connected through a linkage, resulting in an overlapping region on the image plane. This arrangement allows for appropriate movements and rotations. When generating the corresponding bounding box intervals, the algorithm takes into account all the pixels belonging to either the shaft or the metallic clasper, regardless of their presence in the overlapping region ensuring the connectivity between the intervals.

\begin{figure*}[ht]
\centerline{\includegraphics[width=14cm,height=9cm]{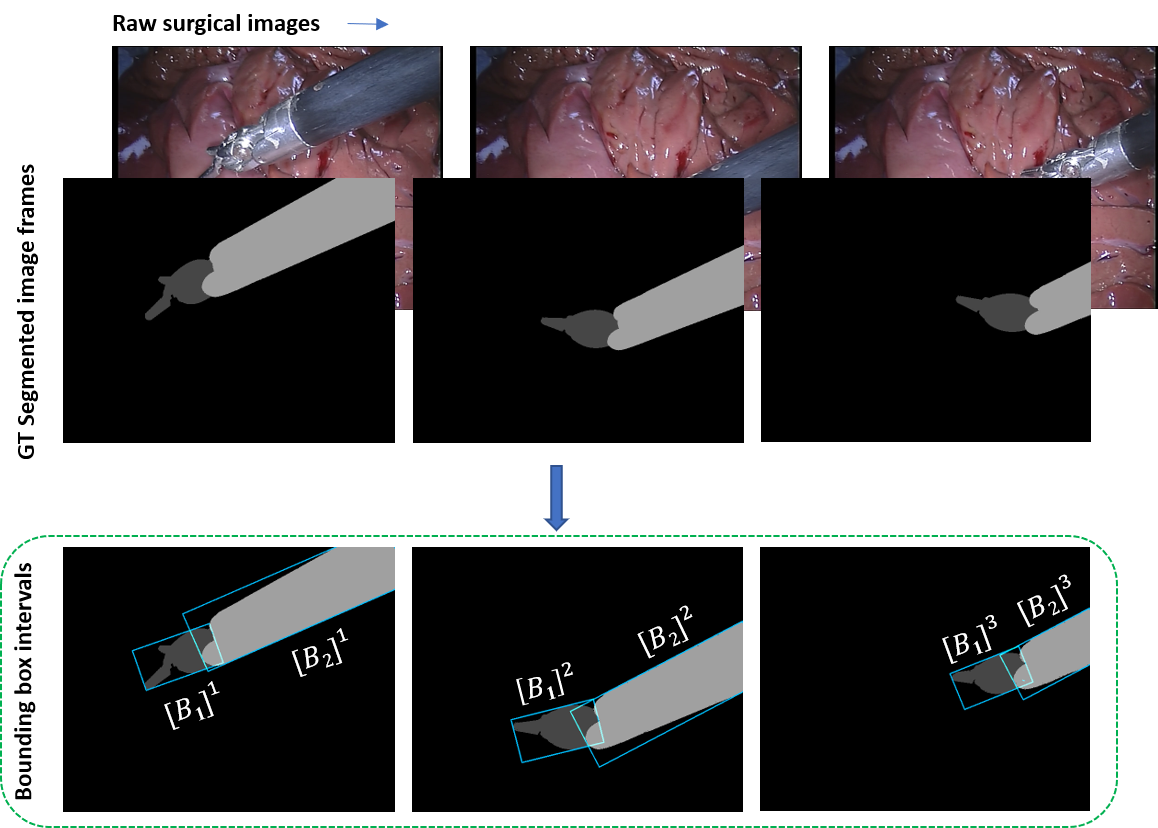}}
\caption{Pictorial representation bounding box intervals formation for shaft
and the metallic clasper}
\label{fig1}
\end{figure*}

\begin{algorithm}

\textbf{Result}:{ $\mathcal{T}_k, \mathcal{R}_k$}
 \textbf{Input:} { $I^n|n=1,2...N$}
 
  \textbf{Initialize:} ${n=1}$
\begin{algorithmic}[1] 
\While{$n <=  N$}
($x_{k,i}^n,y_{k,i}^n$) = $Bound\_box(I^n)$
\For {i = $\{1,2,3,4\}$} 
\For {k = $\{1,2\}$}

     \State Compute:  $A_k^n$ using Eq. (\ref{eq 1})

      \State $S_k^n$ = $A_k^n/A_k^{(n-1)}$

     \State Compute:  $z_k^n$ using Eq. (\ref{eq5})

      \State Compute:  $C_{k,x}^n, C_{k,y}^n, m_{k,i}^n, a_{k,i}^n$ using Eq. (\ref{eq 6}, \ref{eq 4}, \ref{eq 2})

      \State $\Delta x_k^n$ = $C_{k,x}^n - C_{k,x}^{(n-1)}$

     \State  $\Delta y_k^n$ = $C_{k,y}^n - C_{k,y}^{(n-1)}$

      \State $\Delta z_k^n$ = $z_k^n - z_k^{(n-1)}$

     \State  $e_{k,r}^n$ = $a_{k,1}^n + a_{k,4}^n - a_{k,2}^n - a_{k,3}^n$

     \State $\Delta \phi_k$ = $e_{k,r}^n - e_{k,r}^{(n-1)}$

      \State $e_{k,p}^n$ = $a_{k,3}^n + a_{k,4}^n - a_{k,1}^n - a_{k,2}^n$

      \State $\Delta \theta_k$ = $e_{k,p}^n - e_{k,p}^{(n-1)}$

     \State  $e_{k,y}^n$ = $(180 - \sum(tan^{(-1)}(m_{k,i}^n)))/4$

      \State $\Delta \psi_k$ = $e_{k,y}^n - e_{k,y}^{(n-1)}$

     \State  Compute $\mathcal{T}_k$ using Eq. (\ref{eq 7})

      \State Compute $\mathcal{R}_k $ using Eq. (\ref{eq 3})

\EndFor

\EndFor

$n++$
\EndWhile
\end{algorithmic}
\caption{3D movement using 2D boxPoints of clasper and shaft}
\label{algo:3D track}
\end{algorithm}

Following functions are used in the Algorithm~\ref{algo:boundingbox}:
\begin{itemize}
 \item $mask()$: The function detects an object in the segmented image based on the range of pixel intensity values for each $k$. It computes mask ($M_k^n$) in image $I^n$ with pixel intensity values in the specified range. 
 \item $gray()$: This function converts RGB image $O_k^n$ into binary gray scale image $G_k^n$. The image $G_k^n$ has only binary pixel values i.e., either the threshold value or 255. 
 \item $contour( )$: The function computes a curve ($H_k^n$) joining all the continuous points (along the boundary), having same color or intensity. The contour is formed around the shaft or the clasper based on the corresponding specified pixel intensity range. For better accuracy, we applied this function on the binary gray image $G_k^n$. The input parameters are the source image ($G_k^n$), the contour retrieval mode (a mode that returns only extreme outer contours), the contour approximation method (we used chain approximation that stores the pixel coordinates of the boundary of a shape). 

 \item $rect\_bound$():  This function highlights the region of interest as a quadrilateral aligned with the $X-Y$ axis of the image after obtaining contours from an image. The function uses contour list ($H_k^n$) and draws an approximate rectangle around the masked output image ($O_k^n$) with top left corners ($x_k', y_k'$), length $L_k'$ and width $W_k'$ as shown in Fig.~\ref{fig1}

 \item $minAreaRect()$: The function forms a bounding box with minimum area, that also considers the rotation of the detected instrument parts (shaft or the metallic clasper). It returns a Box2D structure ($[B_k]^n$) for given $k$ and $n$ which contains following details - ( center ($x_k,y_k$), (length($l_k$), width($w_k$), angle($\theta_k$) for each contour ($H_k$). 

 \item $boxPoints()$: The function will give the four corners ($x_{k,i}^n,y_{k,i}^n$) of the rotated bounding box for clasper $(k=1)$ and for the shaft $(k=2)$.

\end{itemize}

The image frames ($I^n|n=1,2...N$) in a surgical video with a surgical instrument serve as the input to Algorithm \ref{algo:3D track}. The Algorithm \ref{algo:3D track} is the main algorithm that uses the output from the Algorithm \ref{algo:boundingbox}.
The sequence of operations in Algorithm \ref{algo:3D track} is applied separately for each bounding box points ($(x_{k,i}^n,y_{k,i}^n)$ ( $k=1$for clasper \& $k=2$  for  shaft ) for an image frame $I^n$ obtained using Algorithm \ref{algo:boundingbox} to obtain their 3D motion ($\mathcal{T}_k, \mathcal{R}_k$). 
For each corner box point $i$ and each instrument part $k$, the area ($A_k^n$) is calculated using Eq. \ref{eq 1} and stored in a variable which then computes the scale of change ($S_k^n$). The $z$-coordinates ($z_k^n$) for bounding boxes in each frame is evaluated using Eq.(\ref{eq5}) for which the centroids ($C_{k,x}^n$ and $C_{k,y}^n$) are computed using Eq. (\ref{eq 6}). The slope ($m_{k,i}^n$) between the corner points and  bounding box angles ($a_{k,i}^n$ ) are evaluated  using Eq. (\ref{eq 4}) and Eq. (\ref{eq 2}) respectively. Next, the $\Delta x_k^n$, $\Delta y_k^n$, $\Delta z_k^n$ and $\Delta \phi_k$, $\Delta \theta_k$, $\Delta \psi_k$ are obtained for eventually computing the translation matrix $\mathcal{T}_k$ and rotation matrix $\mathcal{R}_k$.
The transformations obtained for the  shaft and the metallic clasper from frame to frame give a complete 3D track for the surgical instrument.

It is also essential to note, that the moving camera frame only occurs when the surgical view needs to be changed to access a particular area or tissue and this is an intermediate procedure between the main surgical processes. During the surgical procedure, there is minimal to no movement of the camera frame. For incorporating the change of pose due to moving camera frame in practise, change in camera motion can be easily detected by a large change in the tool area/position in subsequent images. This information can be used to initialize the algorithm.  
\section{Simulations and results analysis}
This section illustrates the simulation technique for visualizing results of the proposed approach followed by results to support claims of estimated movements made by the surgical instruments.
\begin{figure}[htbp]
 \centering
\centerline{\includegraphics[width=\linewidth,height=5.5cm]{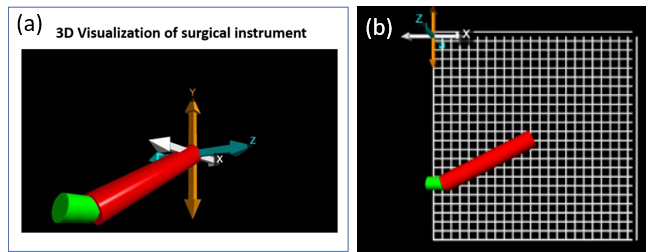}}
\caption{ (a) 3D Visualization and (b) 2D projected frames}
\label{fig2}
\end{figure}

\subsection{3D movement visualization}
The outputs from Algorithm \ref{algo:boundingbox} are generated in the form of corner points of the bounding boxes enclosing the shaft and the metallic clasper. The 2D bounding boxes are simulated using the CV2 python library for a raw surgical video. The simulation results in Fig. \ref{fig1} show that the bounding boxes for the clasper and the shaft are connected via linkage in order to incorporate proper movements and rotations. 
The output of Algorithm \ref{algo:3D track} is the individual movements in the $x$, $y$, and $z$ directions for the clasper and the shaft, respectively, along with their individual roll, pitch, and yaw movements captured through subsequent frames. 
The approach is easily extended to multiple instruments present in the environment based on the range of pixel intensity values of each part (shaft and the metallic clasper) in each instrument. Furthermore, sequential processing of the bounding boxes preserves the connectivity between different parts belonging to a particular instrument. Thus, the individual part movement in 3D is obtained by processing bounding boxes corresponding to each instrument similar to a single instrument.
The objective is to use this output to generate a 3D simulation (refer Fig. \ref{fig2}(a)) that successfully captures and visualizes the robotic arm's movement during endoscopic surgery, hence gives a direct plug-in method for 3D tracking.
A 3D visualization module of Python, Visual Python or vpython (available at glowscript.org), is utilized to simulate the 3D movements of the surgical instrument. 

The clasper and the shaft are simulated as cylindrical objects pertaining to their sizes and orientations. The length of each cylinder is the same as the length of the respective bounding box and the radius of each cylinder is half the width of the respective bounding box while 3D rendering. The base point of the clasper is always attached to the end of the shaft so that both the cylinders representing the shaft and clasper are always connected in the 3D simulation. Using the motion appending function, the $x$, $y$, and $z$ positions of each cylinder’s base are periodically appended by the corresponding output values that are obtained from the algorithm. Moreover, built-in functions in vpython library change the cylinder axes according to the individual roll, pitch, and yaw movements generated. The 3D rendering sampling rate is the same as the frame rate and velocity data. The 3D visualization of the instrument motion is simulated through an interactive screen to view from multiple camera angles for any further analysis.

\subsection{Error Analysis in 2D}

The proposed algorithm is suitable where ground truths for 3D tracking is not available. 
In order to validate our algorithm, we performed error analysis in 2D as well as in 3D. 

To substantiate the proposed 3D visualization, a video is recorded from a fixed camera angle to emulate the raw surgical video.  The corresponding 2D projected image (obtained from a fixed camera angle) of the simulated instrument, a pair of cylinders is shown in Fig.~\ref{fig2} (b). 
The 2D projection of the 3D visualization is on the X-Y plane which was assumed to be the viewing plane of the original raw surgical video captured using the endoscope. This data is then compared to the preexisting data, and the error plot is analyzed. 

Raw surgical videos captured using the endoscope are the only inputs available or rendered in order to perform any intervention. We aimed for the 3D movements of surgical instruments from 2D images to give the surgeons a better prospect. Thus, to improve the efficacy of the 3D simulated results, the 2D raw surgical videos are bench-marked with respect to the emulated 2D video.
\newline
Error in the algorithm is found by reapplying it on the simulation and comparing the obtained trajectory with the original data by measuring the differences in each movement, including the rotation angles. 

\begin{table}
\centering

\caption{Average error in position and rotation between corresponding variables in 2D raw image and 2D synthetic image}
\label{Table 1}
\begin{tabular}{|c|c|c|c|}
\hline 
\textbf{Tool Component}              & \textbf{x  (in pixels)} & \textbf{y  (in pixels)} & \textbf{$\theta$ (in degree)}\\ \hline
Shaft  & 6.472 $\pm$ 2.918 & 8.506 $\pm$ 2.263  & 1.248 $\pm$ 0.734 \\ \hline
Metallic-clasper     & 4.422 $\pm$ 2.856  & 8.456 $\pm$ 4.877 & 1.851 $\pm$ 0.781    \\ 
\hline
\end{tabular}
\end{table}

The average  error (\textit{obtained from dividing the sum of all absolute errors by the number of measurements/frames}) in pixels for $x$ and $y$ movements and in degrees for rotation angle ($\theta$) are obtained between the emulated 2D video corresponding to the 3D movements and the captured video. A tabular representation is given in Table. \ref{Table 1}.  
The results provide evidence of the precise and accurate movements achieved by the instruments with negligible errors.

\subsection{Error Analysis in 3D: Experimental validation}
The proposed algorithms are validated using an experiment conducted for a scaled-up object (a rectangular box with approximate dimensions) executing all the required  movements in all 3 dimensions.
The ground truth for 3D movements is established using a motion capture system that provides motion feedback with very fine resolution.
We captured 10 videos (=10 runs) using a webcam and simultaneously record their ground truth on Vicon software used in motion capture system. The proposed algorithms for 2D tracking followed by 3D tracking are then applied to these videos and the 3D movements for the scaled object (with comparable changes in the image pixels for the corresponding motion between the object and the surgical instrument from EndoVis dataset) are generated in the form of $x$ , $y$ , $z$ movements and roll, pitch and yaw in order to give a complete 3D track for the scaled object.
The relative motion errors in 3D are estimated by comparing the obtained results with the ground truth motion.

\begin{figure*}[htbp]
 \centering
\centerline{\includegraphics[width=\linewidth,height=13cm]{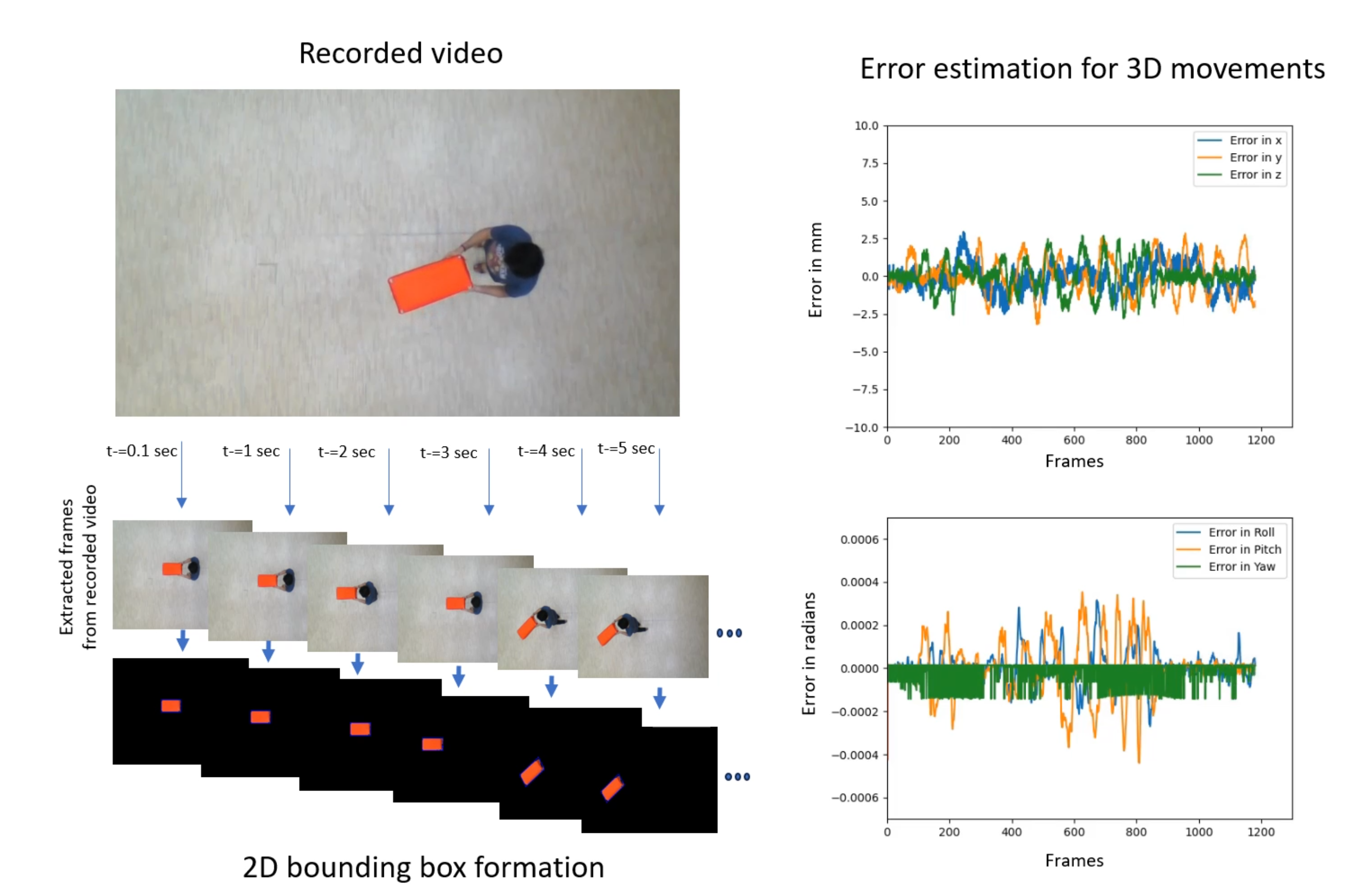}}
\caption{Experimental results from 1 run}
\label{fig7}
\end{figure*}

\begin{table}
\centering

\caption{Average Error (from 10 runs) in position and rotation between corresponding variables in 3D ground truth tracking data collected from motion capture system and 3D tracking obtained using proposed algorithm}
\label{Table 2}
\begin{tabular}{|c|c|c|c|}
\hline 
\textbf{Error variable for the scaled object}              & \textbf{Average Error}     \\ \hline
x (in mm)  &  1.05 $\pm$ 0.05  \\ \hline
y (in mm)   & 0.615 $\pm$ 0.0132 \\ \hline
z (in mm)  &  0.854 $\pm$ 0.0275   \\ \hline
Roll (in rad)  & 1.57 $\times$ $10^{-4}$
 $\pm$ 0.35 $\times$ $10^{-4}$ \\ \hline
Pitch (in rad) & 2.25 $\times$ $10^{-4}$ $\pm$ 0.41 $\times$ $10^{-4}$ \\ \hline
Yaw (in rad) &   1.39 $\times$ $10^{-4}$ $\pm$ 0.65 $\times$ $10^{-4}$ \\ 
\hline
\end{tabular}
\end{table}

A pictorial representation of frames extracted from recorded video and bounding box formation for 2D tracking is shown in left side of Figure \ref{fig7} along with the error estimation in 3D movement is demonstrated in right side of Figure \ref{fig7}. Average Error (from 10 runs) in position and rotation between corresponding variables in 3D ground truth tracking data collected from motion capture system and 3D tracking obtained using proposed algorithm is demonstrated in Table \ref{Table 2}.

\begin{table*}
\centering
\caption{Comparative study on average errors in translational motion}
\label{Table 3}

\begin{tabular}{|cc|ccccc|}
\hline
\multicolumn{2}{|c|}{\multirow{3}{*}{\textbf{Method}}} &
  \multicolumn{5}{c|}{\textbf{Average   Error}} \\ 
\multicolumn{2}{|c|}{} &
  \multicolumn{1}{c|}{\textbf{x   (in mm)}} &
  \multicolumn{1}{c|}{\textbf{y   (in mm)}} &
  \multicolumn{1}{c|}{\textbf{z   (in mm)}} &
  \multicolumn{2}{c|}{\textbf{Translational   motion}} \\  
\multicolumn{2}{|c|}{} &
  \multicolumn{1}{c|}{} &
  \multicolumn{1}{c|}{} &
  \multicolumn{1}{c|}{} &
  \multicolumn{1}{c|}{\textbf{Mean}} &
  \textbf{S.D.} \\ \hline
\multicolumn{1}{|c|}{\multirow{2}{*}{Marker   intensity information \citep{zhang2002application}}} &
  Visual   tracking &
  \multicolumn{1}{c|}{-2.132} &
  \multicolumn{1}{c|}{-12.817} &
  \multicolumn{1}{c|}{-2.999} &
  \multicolumn{1}{c|}{-} &
  - \\ 
\multicolumn{1}{|c|}{} &
  pc   Bird &
  \multicolumn{1}{c|}{-20.82} &
  \multicolumn{1}{c|}{22.452} &
  \multicolumn{1}{c|}{-22.257} &
  \multicolumn{1}{c|}{-} &
  - \\ \hline
\multicolumn{2}{|c|}{Color tape markers (single fisheye camera) \citep{gautier2021real}} &
  \multicolumn{1}{c|}{1.5} &
  \multicolumn{1}{c|}{4} &
  \multicolumn{1}{c|}{3} &
  \multicolumn{1}{c|}{-} &
  - \\ \hline
\multicolumn{2}{|c|}{Monocular  module \citep{chen2023intelligent}} &
  \multicolumn{1}{c|}{-} &
  \multicolumn{1}{c|}{-} &
  \multicolumn{1}{c|}{-} &
  \multicolumn{1}{c|}{2.54} &
  0.22 \\ \hline
\multicolumn{2}{|c|}{Multicamera  module \citep{chen2023intelligent}} &
  \multicolumn{1}{c|}{-} &
  \multicolumn{1}{c|}{-} &
  \multicolumn{1}{c|}{-} &
  \multicolumn{1}{c|}{2.01} &
  0.19 \\ \hline
\multicolumn{2}{|c|}{\begin{tabular}[c]{@{}c@{}}Generalized   hough transform using SIFT \citep{du2016combined}\end{tabular}} &
  \multicolumn{1}{c|}{-} &
  \multicolumn{1}{c|}{-} &
  \multicolumn{1}{c|}{-} &
  \multicolumn{1}{c|}{3.70} &
  2.28 \\ \hline
  \multicolumn{2}{|c|}{Stereoscopic  vision based \citep{duan20113d}} &
  \multicolumn{1}{c|}{1.4} &
  \multicolumn{1}{c|}{0.7} &
  \multicolumn{1}{c|}{0.25} &
  \multicolumn{1}{c|}{0.31} &
  - \\ \hline
\multicolumn{2}{|c|}{\textbf{Proposed   Method}} &
  \multicolumn{1}{c|}{\textbf{1.05}} &
  \multicolumn{1}{c|}{\textbf{0.615}} &
  \multicolumn{1}{c|}{\textbf{0.854}} &
  \multicolumn{1}{c|}{\textbf{0.3989}} &
  \textbf{0.303} \\ \hline
\end{tabular}
\end{table*}

\subsection{Comparative study}

In previous works on tracking algorithms, average errors for $x$, $y$ and $z$ movements (translational motion) have been reported using different setups mimicking surgical instrument. Reporting of the quantitative rotational motion error are missing to the best of our knowledge. 
We present a comparative study of the proposed work with minimum error reported in literature (in Table \ref{Table 3}). 
The method utilizing intensity information of the image for the marker localization \citep{zhang2002application} gives huge errors for 3D tracking using pc Bird, that reduces to a significant amount using visual tracking. Our proposed method reduces this error further by 50.8\%, 94.99\% and 71.52\% for $x$, $y$ and $z$ movement respectively computed from data reported in Table  \ref{Table 3}.
The percentage reduction in errors by our method for x,y and z movements are 30\%, 84.63\% and 71.53\% for a similar method that uses color tape markers to capture 3D movements using a single fisheye camera \citep{gautier2021real}. 
The average translational error is reduced by more than 80\% using the proposed method as compared to methods that uses monocular module \citep{chen2023intelligent}, multicamera module and a combined 2D3D method in which generalized hough transform is utilized for SIFT feature extraction \citep{du2016combined}.
Thus, it is observed that the proposed method reduces the individual and combined translational average error by a decent margin as compared to existing state-of-the-art methods and produces the results comparable to the stereoscopic vision based method \citep{duan20113d}.

Moreover, the tracking algorithms run on a computing machine with i7 compatible with CPU/GPU. The execution time for processing 1270 image frames is $\approx$ 20 ms for the complete 3D tracking with segmented maps available a-priori and consumes only 46$\%$ of CPU memory that makes the method computationally inexpensive.

\section{Conclusion and Future work}
In this paper, we present a novel 2D tracker for the segmentation maps of surgical instruments, which combines with a 3D tracking algorithm to robustly estimate the full 3D pose of instruments in minimally invasive surgery. The proposed plug-in approach produces tracking results for any number of segmented instruments present in the surgical environment without any physical intervention, making it likely to succeed in market dissemination. 
The proposed method considers changing intervals of the surgical instruments (for the connected set of the shaft and the metallic clasper) while moving in a surgical environment cluttered with obstacles (sensitive tissues, veins) and uses lower dimensional results to track surgical instruments in higher dimensions. Synthesized and experimental results in 2D and 3D motion estimates show that errors are negligible for the method to be used for dataset labelling and motion tracking of instrument for futuristic applications.
Due to the use of simple computations, future developments in this method will focus on the real-time implementation of a 3D tracking system. 
It is interesting to note that connected parts angular motion is independent of each other, and a methodology to extract 3D independent motions of each part from their 2D motion in image frames has multiple solutions giving an opportunity for future investigations. Moreover, we will focus on developing algorithms that effectively track the dynamic reference frame's motion and consistently transform tool poses to the desired fixed reference frame, ensuring robust and reliable results.

\section*{Acknowledgments}
Financial support for this study was provided by Prime Minister's Research Fellows (PMRF) scheme, India (PMRF Id no. 1300229 dated May 2019) for pursuing research in higher educational institutions in India. \url{https://www.pmrf.in/}.
Thanks are also due to Indian Institute of Technology, Bombay, India.

\section*{Statements and Declarations}

\subsection*{Funding:} 
This work was supported by Prime Minister's Research Fellows (PMRF) scheme. PMRF Id no. 1300229 awarded to Shubhangi Nema.
\subsection*{Conflict of interest disclosure:}
The authors declare that they have no conflict of interest.

\subsection*{Ethical approval:} This article does not contain any studies with human participants or animals performed by any of the authors.

\subsection*{Authors’ contributions:}
All authors contributed to the study conception and design. Finding research gap, idea generation, data collection, experiments, analysis, and interpretation of results were performed by Shubhangi Nema and Abhishek Mathur. Leena Vachhani advised on the findings of this work, reviewed the methods, and agreed on the novelty of the article before submission. The first draft of the manuscript was written by Shubhangi Nema and all authors commented on previous versions of the manuscript. All authors read and approved the final manuscript.

\subsection*{Data availability:}  
The author confirms that all data generated or analysed during this study are included in this published article.

\bibliography{ref}

\begin{thebibliography}{10}
\providecommand \doibase [0]{http://dx.doi.org/}%

\bibitem{R1}
Galuret S, Vall{\'e}e N, Tronchot A, Thomazeau H, Jannin P, Huaulm{\'e} A. Gaze behavior is related to objective technical skills assessment during virtual reality simulator-based surgical training: a proof of concept. {\it International Journal of Computer Assisted Radiology and Surgery} 2023\string: 1--9.

\bibitem{R2}
Basdogan C, Sedef M, Harders M, Wesarg S. {VR}-based simulators for training in minimally invasive surgery. {\it IEEE Computer Graphics and Applications} 2007\string; 27(2)\string: 54--66.

\bibitem{R3}
Suh I, Mukherjee M, Oleynikov D, Siu KC. Training program for fundamental surgical skill in robotic laparoscopic surgery. {\it The International Journal of Medical Robotics and Computer Assisted Surgery} 2011\string; 7(3)\string: 327--333.

\bibitem{beulens2020analysis}
Beulens AJ, Namba HF, Brinkman WM, et al. Analysis of the video motion tracking system “Kinovea” to assess surgical movements during robot-assisted radical prostatectomy. {\it The International Journal of Medical Robotics and Computer Assisted Surgery} 2020\string; 16(2)\string: e2090.

\bibitem{nema2022surgical}
Nema S, Vachhani L. Surgical instrument detection and tracking technologies: Automating dataset labeling for surgical skill assessment. {\it Frontiers in Robotics and AI} 2010\string: 308.

\bibitem{dataset}
Speidel S. Endovis sub-challenge: Instrument segmentation and tracking.. \url{https://endovissub-instrument.grand-challenge.org/Data/};  2012-22.

\bibitem{Pakhomov}
Pakhomov D, Premachandran V, Allan M, Azizian M, Navab N. Deep residual learning for instrument segmentation in robotic surgery. In:  {\it International Workshop on Machine Learning in Medical Imaging}Springer. ; 2019\string: 566--573.

\bibitem{Shvets}
Shvets AA, Rakhlin A, Kalinin AA, Iglovikov VI. Automatic instrument segmentation in robot-assisted surgery using deep learning. In:  {\it 2018 17th IEEE International Conference on Machine Learning and Applications (ICMLA)}IEEE. ; 2018\string: 624--628.

\bibitem{nema2023unpaired}
Nema S, Vachhani L. Unpaired deep adversarial learning for multi-class segmentation of instruments in robot-assisted surgical videos. {\it The International Journal of Medical Robotics and Computer Assisted Surgery} 2010\string: e2514.

\bibitem{zhang2002application}
Zhang X, Payandeh S. Application of visual tracking for robot-assisted laparoscopic surgery. {\it Journal of Robotic systems} 2002\string; 19(7)\string: 315--328.

\bibitem{R4}
Schulze F, B{\"u}hler K, Neubauer A, Kanitsar A, Holton L, Wolfsberger S. Intra-operative virtual endoscopy for image guided endonasal transsphenoidal pituitary surgery. {\it International journal of computer assisted radiology and surgery} 2010\string; 5(2)\string: 143--154.

\bibitem{R5}
Choi C, Kim J, Han H, Ahn B, Kim J. Graphic and haptic modelling of the oesophagus for {VR}-based medical simulation. {\it The International Journal of Medical Robotics and Computer Assisted Surgery} 2009\string; 5(3)\string: 257--266.

\bibitem{zhang2017real}
Zhang L, Ye M, Chan PL, Yang GZ. Real-time surgical tool tracking and pose estimation using a hybrid cylindrical marker. {\it International journal of computer assisted radiology and surgery} 2017\string; 12\string: 921--930.

\bibitem{R6}
Krieg J. Motion tracking: Polhemus technology. {\it Virtual Reality Systems} 1993\string; 1(1)\string: 32--36.

\bibitem{R7}
Trejos AL, Patel RV, Naish MD, Schlachta CM. Design of a sensorized instrument for skills assessment and training in minimally invasive surgery. In:  {\it 2008 2nd IEEE RAS \& EMBS International Conference on Biomedical Robotics and Biomechatronics}IEEE. ; 2008\string: 965--970.

\bibitem{R8}
Yamaguchi S, Yoshida D, Kenmotsu H, et al. Objective assessment of laparoscopic suturing skills using a motion-tracking system. {\it Surgical endoscopy} 2011\string; 25(3)\string: 771--775.

\bibitem{R9}
Stoll J, Novotny P, Howe R, Dupont P. Real-time {3D} ultrasound-based servoing of a surgical instrument. In:  {\it Proceedings 2006 IEEE International Conference on Robotics and Automation, 2006. ICRA 2006.}IEEE. ; 2006\string: 613--618.

\bibitem{R11}
Halic T, Kockara S, Bayrak C, Rowe R. Mixed reality simulation of rasping procedure in artificial cervical disc replacement ({ACDR}) surgery. In:  {\it BMC bioinformatics}. 11. Springer. ; 2010\string: 1--17.

\bibitem{seslija2008feasibility}
Seslija P, Habets DF, Peters TM, Holdsworth DW. Feasibility of {3D} tracking of surgical tools using 2D single plane X-ray projections. In:  {\it Medical Imaging 2008: Visualization, Image-Guided Procedures, and Modeling}. 6918. SPIE. ; 2008\string: 212--221.

\bibitem{duan20113d}
Duan Z, Yuan Z, Liao X, Si W, Zhao J. 3D tracking and positioning of surgical instruments in virtual surgery simulation.. {\it Journal of multimedia} 2011\string; 6(6).

\bibitem{gautier2021real}
Gautier B, Tugal H, Tang B, Nabi G, Erden MS. Real-time 3D tracking of laparoscopy training instruments for assessment and feedback. {\it Frontiers in Robotics and AI} 2021\string; 8\string: 751741.

\bibitem{chen2023intelligent}
Chen L, Ma L, Zhang F, Yang X, Sun L. An intelligent tracking system for surgical instruments in complex surgical environment. {\it Expert Systems with Applications} 2023\string: 120743.

\bibitem{jiang2015video}
Jiang X, Zheng B, Atkins MS. Video processing to locate the tooltip position in surgical eye--hand coordination tasks. {\it Surgical innovation} 2015\string; 22(3)\string: 285--293.

\bibitem{allen2011visual}
Allen BF, Kasper F, Nataneli G, Dutson E, Faloutsos P. Visual tracking of laparoscopic instruments in standard training environments. In: IOS Press.  2011 (pp. 11--17).

\bibitem{du2016combined}
Du X, Allan M, Dore A, et al. Combined {2D} and {3D} tracking of surgical instruments for minimally invasive and robotic-assisted surgery. {\it International journal of computer assisted radiology and surgery} 2016\string; 11(6)\string: 1109--1119.

\bibitem{garcia2016real}
Garc{\'\i}a-Peraza-Herrera LC, Li W, Gruijthuijsen C, et al. Real-time segmentation of non-rigid surgical tools based on deep learning and tracking. In:  {\it International Workshop on Computer-Assisted and Robotic Endoscopy}Springer. ; 2016\string: 84--95.

\bibitem{cheng2021deep}
Cheng T, Li W, Ng WY, et al. Deep learning assisted robotic magnetic anchored and guided endoscope for real-time instrument tracking. {\it IEEE Robotics and Automation Letters} 2021\string; 6(2)\string: 3979--3986.

\bibitem{ronneberger2015u}
Ronneberger O, Fischer P, Brox T. U-net: Convolutional networks for biomedical image segmentation. In:  {\it International Conference on Medical image computing and computer-assisted intervention}Springer. ; 2015\string: 234--241.

\bibitem{simonyan2014very}
Simonyan K, Zisserman A. Very deep convolutional networks for large-scale image recognition. {\it arXiv preprint arXiv:1409.1556} 2014.

\bibitem{nema2021safe}
Nema S, Vachhani L. Safe and Fast Path Planner for Minimally Invasive Surgery. In:  {\it 2021 IEEE/RSJ International Conference on Intelligent Robots and Systems (IROS)}IEEE. ; 2021\string: 2549--2554.

\end{thebibliography}

\end{document}